\pdfoutput=1

\documentclass[11pt]{article}

\usepackage[final]{acl}

\usepackage{times}
\usepackage{latexsym}

\usepackage[T1]{fontenc}

\usepackage[utf8]{inputenc}

\usepackage{microtype}

\usepackage{inconsolata}

\usepackage{graphicx}

\usepackage{amsmath}
\usepackage{amsthm}
\usepackage{amssymb}
\usepackage{graphicx}
\usepackage{comment}
\usepackage{multirow}
\usepackage{enumitem}
\usepackage{booktabs}
\usepackage{tikz}
\usepackage{linguex}
\usepackage{array}
\usepackage{float}
\usepackage{algorithm}
\usepackage{algpseudocode}
\usepackage{algcompatible}
\usepackage{hyperref}

\theoremstyle{definition}

\newcolumntype{S}{>{\small}c}

\title{\textsc{MGen}: Millions of Naturally Occurring Generics in Context}

\author{Gustavo Cilleruelo Calderón \phantom{aaa} Emily Allaway \phantom{aaa} Barry Haddow \phantom{aaa} Alexandra Birch \\
School of Informatics, University of Edinburgh\\
\texttt{\small g.cilleruelo-calderon@sms.ed.ac.uk \phantom{aaa} \{emily.allaway, bhaddow, a.birch\}@ed.ac.uk}}

\begin{document}
\maketitle
\begin{abstract}
\textsc{MGen} is a dataset of over $4$ million naturally occurring generic and quantified sentences extracted from diverse textual sources. 
Sentences in the dataset have long context documents, corresponding to websites and academic papers, and cover 11 different quantifiers. 
We analyze the features of generics sentences in the dataset, with interesting insights: generics can be long sentences (averaging over 16 words) and speakers often use them to express generalisations about people.

\textsc{MGen} is the biggest and most diverse dataset of naturally occurring generic sentences, opening the door to large-scale computational research on genericity. It is publicly available at \href{https://gustavocilleruelo.com/mgen}{\tt gustavocilleruelo.com/mgen}.
\end{abstract}

\section{Introduction}

\begin{table*}[t]
  \small
  \centering
  \begin{tabular}{p{2cm} p{10cm} }
  \toprule
    \multicolumn{1}{l}{{\bf Source}} & \multicolumn{1}{l}{{\bf Sentence}}\\
  \midrule

  RefinedWeb&Soybeans contain an inhibitor of trypsin, an enzyme important for digestion, but it can be destroyed by cooking.\\[0.6cm]

  SlimPajama&Cucumbers are high in an antioxidant called beta-carotene, which your body turns into vitamin A. May ease muscle cramps.  \\[0.6cm]

  The Pile&Starving people grab the bread first and run with it.\\[0.2cm]

  arXiv&Colexification networks encode affective meaning.\\[0.2cm]

  peS2o&Car seats save lives.\\

  \bottomrule

  \end{tabular}
  \caption{Examples of generic sentences from the different sources of \textsc{MGen}. More examples in Appendix \ref{apx:annotations}.}
  \label{tab:congen_examples}
\end{table*}

Generics are sentences that express generalisations without making use of explicit quantifiers. Examples of generics are \textit{ravens are black} or \textit{ticks carry lyme disease}. 

Several features of generics make them difficult to account for semantically \citep{carlson1995thegenericbook}: they are permissive to exceptions (\textit{ravens are black} is acceptable even if albino ravens exist) and the quantifications they convey have paradoxical dynamics \citep{generics_08_leslie}. 
If we paraphrase the previous generics as explicitly quantified, we would have \textit{most ravens are black} but \textit{few ticks carry lyme disease}: the same linguistic structure conveys generalisations at opposite ends of the quantification spectrum.



In this work, we introduce \textsc{MGen}, a dataset designed to provide a solid foundation for research on generic sentences in English. \textsc{MGen} has $4.1$ million samples, with over $3$ million generics and $1$ million explicitly quantified sentences with 11 different quantifiers. All sentences are naturally occurring and include the context document in which they originally appear. 

To motivate the design of \textsc{MGen}, we conduct an extensive review of datasets of generic sentences and argue that existing datasets have many shortfalls: they are either small, rely on synthetic samples or have no context, despite theoretical works showing the importance of context for the semantics of generics \citep{sterken2015context,gencontext_23_almotahari}.






In order to mine generic sentences from massive corpora, we introduce a two-step pipeline: a syntactic filter detects bare plurals (this is the most common syntax of the subject for generics, see \S\ref{sec:background})
with the required verb features 
and then a binary classifier labels them as generic or not. We apply this pipeline to a subset of the \textsc{Zyda} \citep{tokpanov2024zyda} dataset (a language model pre-training corpus) to collect a diverse  and accurate (as per human annotators) dataset of generic sentences. 

We analyze the corpus-level characteristics of \textsc{MGen} and find that its generic sentences are longer than those usually considered in the literature, where running examples are much shorter than the average  16.65 words in our dataset. Analysing the word frequencies of our dataset, we find that speakers use generics most often to generalize about \textit{people}.


Our contributions are: (i) \textsc{MGen}, the largest dataset of naturally occurring 
generics in context, (ii) a pipeline for the extraction bare plural generics from textual sources, (iii) a review of existing datasets of generics and (iv) a preliminary corpus-level analysis of the characteristics of generic sentences.

\section{Background: generics \& quantifiers}
\label{sec:background}

Generics have kind terms in their subject position (i.e. words or phrases used to categorize or label groups of entities) and their verbs are inflected for third person plural present indicative. They are used either to make claims about those kinds (\textit{dinosaurs are extinct}) or to attribute properties to individuals in those kinds (\textit{beetles have protective wing covers}). 

Following most of the linguistics and philosophy of language literature, we consider only \textit{bare plural} generics \citep{carlson1995thegenericbook, leslie2007thesis}. Bare plurals have noun phrases in
plural form without a definite or indefinite article\footnote{\textit{Tigers have stripes} is a bare plural generic, which can also be expressed in English with the definite (\textit{the tiger has stripes}) or indefinite (\textit{a tiger has stripes}) articles.}. Throughout the paper, we will use \textit{bare plural sentence} to refer to sentences with the syntax of a bare plural generic (i.e. with the same inflection of the verb), even if those sentences are not generics. 

The standard view in linguistics is that generics are quantificational: there is an unpronounced operator \textsc{gen} that takes a role similar to adverbial quantifiers in the logical form of the sentence \citep{lewis1975adverbs,carlson1977disseration,carlson1995thegenericbook,cohen1999thinkgeneric,sterken2015context,nickel2016between, kirkpatrick2024quantificational}.
In contrast, recent influential accounts of generics have been non-quantificational:
\citet{generics_08_leslie} gives generics the privileged role of expressing default or primitive generalisations.

The rich landscape of theories of generics, as well as their far-reaching implications into fundamental aspects of human cognition, has made the study of generic sentences a highly debated topic in recent years ~\citep[e.g.,][]{cohen1999probability,goodman2018generalization,stovall2019tickets,nguyen2020radical,bosse2021nonspecific,weak_22_almotahari,kirkpatrick2023dynamics,neufeld2025giving}

In the field of natural language processing, recent works study how language models deal with aspects of genericity such as exceptions, property inheritance \citep{allaway2024exemplars_llms} and quantification \citep{ralethe-buys_2022_generic, collacciani2024quantifying}. \citet{cilleruelo2025generics} uses language models to study the semantics of generic sentences, such as their implicit quantification.

\section{Related work: datasets of generics}
\label{sec:related_work}

\begin{table*}
  \small
  \centering
  \begin{tabular}{l r r r l}
  \toprule
   {\bf Dataset} & {\bf Scale} & {\bf Quantifiers} & {\bf Context} & {\bf Sources} \\
  \midrule
  \textsc{MGen} (Ours) & $4.1M$ & Yes (11) & Yes & Natural (\textsc{Zyda}) \\[0.2cm]
  
  \textsc{GenericsKB-Best} & $1M$ & Yes (13) & Yes & Natural (Waterloo, SimpleWiki, ARC) \\ \citep{Bhakthavatsalam_2020_genericskb} & & & & Synthetic (WordNet, ConceptNet, TupleKB)\\[0.2cm]
  \textsc{ConGen} \citep{cilleruelo2025generics} & 2872 & Yes (3) & Yes & Natural (\textsc{Dolma}) \\[0.2cm]
  \textsc{Gen-A-Tomic}  & $>8M$ & Yes (3) & No & Synthetic (\textsc{Gpt2-XL} with \textsc{I2D2}) \\
  \citep{bhagavatula2023i2d2} & & & & \\[0.2cm]
  Animal generics  & 75,002 & No & No & Mixed (\textsc{GenericsKB}) \\
  \citep{ralethe-buys_2022_generic}& & & & \\[0.2cm]
  \textsc{Exemplars} (generics)  & $16,655$ & No & No & Mixed (\textsc{Gen-A-Tomic}, Animal generics)\\
  \citep{allaway2024exemplars_llms} & & & & \\[0.2cm]
  Dataset in \citep{collacciani2024quantifying} & 1837 & Yes (5) & No & Synthetic (human annotations) \\[0.2cm]
  Norwegian generics  & $170$ & No & Yes & Natural (encyclopedia entries) \\
  \citep{KurekPrzybilski2022GenericsAA} & & & & \\
  \bottomrule
  \end{tabular}
  \caption{Comparison between existing datasets of generic sentences. \textsc{MGen} is comparable in size with synthetic datasets but is comprised of naturally occurring sentences in context.}
  \label{tab:dataset_comparison}
\end{table*}


Several datasets exist that specifically target generics. 
We compare these datasets across four dimensions (Table \ref{tab:dataset_comparison}): total samples, quantified sentences, context and origin (natural or synthetic). 

We consider \textit{natural} sentences to be only those that have been extracted from human-written sources and \textit{synthetic} those have been either generated by language models, built with rule-based methods or constructed by researchers or annotators.
We also include quantified sentences as a requirement for datasets of generics as these are a key contrast class. Similarly, context plays an important role on the semantics of generics.

\textsc{GenericsKB} \citep{Bhakthavatsalam_2020_genericskb} is a dataset that is composed of both naturally occurring generic and quantified sentences in context and synthetic examples derived from knowledge bases. 

To source the naturally occurring samples, $3.5M$ candidate sentences are extracted from different corpora (Wikipedia, ARC and Waterloo) through 27 hand-crafted lexico-semantic rules. A subset of those are manually annotated and used to train a \textsc{Bert}-based binary classifier (generic and not generic). 

This classifier is used to score the $3.5M$ candidate sentences to curate \textsc{GenericsKB-Best}: a collection of the best-scoring naturally occurring sentences ($N=774,621$) augmented with synthetic generics derived from knowledge bases ($N=246,247$). Some sentences are quantified with \textit{all}, \textit{most}, \textit{some}, \textit{many}, \textit{every}, \textit{much}, \textit{more}, \textit{often}, \textit{usually}, \textit{always}, \textit{sometimes}, \textit{frequently}. 

\citet{cilleruelo2025generics} introduce \textsc{ConGen}, a collection of 2873 naturally occurring generic and quantified sentences in context. Because the dataset is manually curated, it is small and only contains data for 3 quantifiers (\textit{all}, \textit{most} and \textit{some}).



The biggest dataset of synthetic generics is the \textsc{Gen-A-Tomic} corpus \citep{bhagavatula2023i2d2}. Sentences in \textsc{Gen-A-Tomic} are generated by \textsc{Gpt2-xl}~\citep{radford2019gpt2} through knowledge distillation with self-imitation algorithm. Although \textsc{Gen-A-Tomic} has over $8$ million utterances, because they are generated with a small language model, these are not in context and the only quantifiers included are \textit{generally}, \textit{typically} and \textit{usually}.

\citet{ralethe-buys_2022_generic} select generics and quantified sentences from \textsc{GenericsKB} by filtering for animals, curating a subset of 75,002 generics.
This collection of animal generics is combined with examples from \textsc{Gen-A-Tomic} to create datasets of synthetic generics exemplars (i.e. cases where the generic does and does not hold)~\citep{allaway2023penguins, allaway2024exemplars_llms}, which contain generic sentences, as well as their derived exemplars.

To conduct experiments on language models, \citet{collacciani2024quantifying} collect 1873 sentences from three sources, all crafted either by researchers or annotators \citep{herbelot2016many, urbach2010quantifiers,misra2023comps}. Sentences in this dataset are extremely short (average length is $3.73\pm1.03$, median is $3$) and all are annotated with a quantifier (\textit{all}, \textit{most}, \textit{some}, \textit{few}, \textit{no}).

All datasets considered so far, as well as \textsc{MGen}, are in English. In Norweigan, \citet{KurekPrzybilski2022GenericsAA} manually extract 170 generics in context from encyclopedic texts.

Table \ref{tab:dataset_comparison} compares the reviewed datasets of generic sentences in terms of total samples, inclusion of quantified sentences, context for the utterances and data origin.
Our dataset, \textsc{MGen}, has the scale of \textsc{GenericsKB} and \textsc{Gen-A-Tomic}, but without the need of synthetic examples (whether generated or constructed from knowledge bases) and includes context documents for all generic as well as quantified utterances. 

\section{Methodology}
\label{sec:methodology}

This section details the construction of the \textsc{MGen} dataset. We first describe the high-level objectives for the creation of the dataset, based on the generics literature and the shortcomings of existing datasets. Then, we detail the extraction of generics and quantified sentences at scale from a large corpus by leveraging syntactic (\S\ref{sec:syntactic_filter}) and semantic (\S\ref{sec:semantic_filter}) characteristics of generics.

\subsection{Design choices}

\textsc{MGen} is built to include a massive, diverse amount of naturally occurring generic sentences with their respective contexts. In this section we go over the principles that guide the construction of the dataset.

\paragraph{Naturally occurring.} We focus on naturally occurring generic sentences, as it would be hard to assess the acceptability of synthetic samples 
without assuming a theory of generics or conducting extensive human annotation studies, since the semantics of generics are not well understood (\S\ref{sec:background}).

\paragraph{Context.} Many works argue that the context radically affects what generic sentences express, for example, in terms of both quantificational strength and flavor \citep{sterken2015context, gencontext_23_almotahari}. To mine generic sentences, we choose a corpus structured in documents (more details in \S\ref{sec:data_sources}) and keep the full context document of each sample.

\paragraph{Bare plurals.} We focus on generics that are bare plurals (\S\ref{sec:background}) and only at the beginning of a sentence. This makes detection at scale more tractable, by, for example, omitting nested generics in \textit{that} clauses (e.g. \textit{she maintains that the belief that technology improves education is widely accepted}). 

\paragraph{Quantifiers.} Generics and quantified sentences are closely related, as both are used to express generalisations.
We collect quantified sentences with the following structures: \textit{quantifier $+$ bare plural sentence}, \textit{bare plural noun phrase $+$ quantifier $+$ verb} or \textit{bare plural noun phrase $+$ verb $+$ quantifier}. 
We consider the following 11 quantifiers: \textit{all}, \textit{most}, \textit{many}, \textit{some}, \textit{few}, \textit{no}, \textit{often}, \textit{generally}, \textit{typically}, \textit{usually}, \textit{normally}. 




\subsection{Data sources}
\label{sec:data_sources}

Training language models requires large collections of clean textual data, which can also be used for data mining. We use \textsc{Zyda} \citep{tokpanov2024zyda}, an open-source dataset built by collecting text from different high-quality sources and performing uniform filtering and deduplication. 
We run our generic extraction pipeline on the following components of \textsc{Zyda} (Appendix \ref{apx:mgen}; Table \ref{tab:zyda_components}): RefinedWeb \citep{penedo2023refinedweb}, SlimPajama \citep{cerebras2023slimpajama}, the Pile \citep{gao2021thepile}, peS2o \citep{peS2o} and arXiv \citep{kenney2023arxiv}. 

RefinedWeb, SlimPajama and The Pile 
primarily consist of data scraped from the web, while the much smaller peS2o and arXiv are composed of academic publications.

\subsection{Generic sentence extraction}

\textsc{Zyda} is structured in documents: roughly the text in a website, a scientific article or similar. 
Each document is first split into sentences (\texttt{blingfire}\footnote{\href{https://github.com/microsoft/BlingFire}{https://github.com/microsoft/BlingFire}}).  Then, a lightweight syntactic filtering step selects sentences where either (i) the first word is one of the quantifiers of interest, or (ii) there is a \textit{plural noun} in the first 4 words of the sentence (\texttt{flair} \citep{akbik2019flair}). 

These candidates are then run through two filtering steps: a syntactic one that ensures these are bare plurals with verbs inflected for third person present indicative and a semantic one, that filters for sentences that express generalizations.
This latter step is necessary as the bare plural generic syntactic construction can also have existential readings, where the subject refers to specific instances instead of to a kind in general, e.g. \textit{tigers are in the front lawn} or \textit{blue arrows indicate acceleration} (also see Appendix \ref{apx:annotations}; Table \ref{tab:existentials}).

We detail the construction of each filtering step in \S\ref{sec:syntactic_filter} and \S\ref{sec:semantic_filter}  respectively.



\subsection{Syntactic filtering (bare plurals)}
\label{sec:syntactic_filter}

The syntactic filtering step in the pipeline receives candidate sentences with plural nouns in the early words and performs a more in-depth dependency analysis to select only bare plural sentences. 

The part-of-speech and dependency parsing of the sentence is conducted with the \texttt{stanza} python library \citep{stanza}. After parsing the sentences, we keep those that meet the following three conditions:

\begin{enumerate}
  \item The nominal subject is a plural noun or a plural proper noun (\texttt{nsubj} or \texttt{nsubj:pass} in the case of passives). 
  \item The root of the nominal subject is a verb or an auxiliary (\texttt{VERB} or \texttt{AUX}). If there is a copula (\texttt{cop}) or a passive (\texttt{aux:pass}), take that as the verb. 
  \item The verb has present tense, indicative mood, plural number and third person.
\end{enumerate}

\subsection{Semantic filtering (genericity)}
\label{sec:semantic_filter}

The syntactic filtering step yields bare plural candidate sentences, but these include noisy and non-generic samples.
To get high quality generics from these candidates, we apply a further step in which a binary classifier scores whether the bare plurals are generic or not.

This classifier is designed to filter out: (i) sentences that although they may contain a generic it is not at the beginning\footnote{A common occurrence are titles of paragraphs or sections that get parsed at the beginning of the sentence, for example: \textit{Gaussian Mixture Models  Gaussian mixture models are formed by combining multivariate normal \dots}. Note how the title (\textit{Gaussian Mixture Models}) makes it so that the generic is not at the beginning.}, (ii) sentences that are ungrammatical or noisy and (\textit{iii}) bare plurals that have existential (non-generic) readings (Table \ref{tab:existentials}).

We use a \textsc{RoBERTa} model \citep{liu2019roberta} as the architecture for the classifer, which we train on a small collection of generics and non-generic bare plurals. 
The generics are sampled from \textsc{GenericsKB-Best} and the non-generics are generated by \textsc{GPT-4} \citep{openai2024gpt4technicalreport}, by iteratively finding missclassified examples to make the training data more robust. The classifier achieves over $0.97$ F-$1$ score in a test set based on \textsc{ConGen} and synthetic non-generic bare plurals. More details on classifier training and evaluation are found in Appendix \ref{apx:roberta}.

In the case of sentences that start with a quantifier, which are not bare plurals and are outside of the training distribution of the generics classifier, we remove the quantifier word and calculate the score of the resulting bare plural. This ensures that we pick out quantified sentences that are comparable to generics in terms of being generalizations as opposed to existential.  We want to keep in the dataset sentences like \textit{all tigers have stripes} but not \textit{all tigers in the cage are male}.

Some quantified sentences begin with a bare plural rather than a quantifier (e.g. \textit{tigers are normally striped}). For these sentences, we check if there is an adverbial quantifier that has as syntactic head the root of the sentence, and label them with the corresponding quantifier (if the quantifier is not in the main clause, the sentence is labeled as generic).

We include sentences that receive a genericity classifier score $0.8$ or greater for the \textsc{MGen} dataset. This value is chosen by manual inspection of the data. The full unfiltered bare plurals data is also made publicly available.

\begin{table}
  \centering\small
  \begin{tabular}{l r r}
  \toprule
      
      & {\bf Candidates} & {\bf Generalizations } \\
      \midrule
      \textsc{gen} & $14,303,840$ & $3,183,293$ \\
      All       & $502,629$ & $82,752$ \\
      Most      & $332,698$ & $173,021$ \\
      Many      & $389,606$ & $188,419$ \\
      Some      & $547,308$ & $225,171$ \\
      Few       & $22,164$ & $8,085$ \\
      No        & $47,146$ & $4,121$ \\
      Generally & $116,901$ & $53,015$ \\
      Typically & $124,522$ & $53,046$ \\
      Often     & $253,306$ & $107,926$ \\
      Usually   & $138,207$ & $59,148$\\
      Normally  & $19,969$ & $8,763$ \\
      \midrule
      TOTAL & $16,771,049$ & $4,146,760$ \\
      \bottomrule
  \end{tabular}
  \caption{Number of generics and quantified sentences after syntactic (candidates) and semantic (generalizations) filtering during the construction of \textsc{MGen}.}
  \label{tab:updated_combined}
\end{table}

\begin{table*}[h]
  \small
  \centering
  \begin{tabular}{p{8cm} l l c}
  \toprule
    \textbf{Text} & \textbf{Label 1} & \textbf{Label 2} & \textbf{Score} \\
    \midrule
    Puppets are fun to include too.&Particular&Unclear&0.86 \\[0.2cm]
    First thoughts are proverbially the best; at all events, they are the bravest.&Unclear&Generic&0.96\\[0.6cm]
    Pumps are used to circulate the water through collectors and into your water tanks.&Particular&Generic&0.97\\[0.6cm]
    Players get sets by asking another player for a specific card.&Generic&Particular&0.82\\
    
      \bottomrule
  \end{tabular}
  \caption{Examples of annotator disagreements with classifier scores.}
  \label{tab:annotation_disagreements}
\end{table*}

\section{\textsc{MGen}: Statistics \& Analysis}
\label{sec:analysis}

In this section we summarize the statistics of the \textsc{MGen} dataset (\S\ref{sec:statistics}) and present two quality analyses: human annotation to asses the genericity of the collected sentences (\S\ref{sec:genericity}) and a comparison in terms of diversity with existing datasets (\S\ref{sec:diversity}).

\subsection{Statistics}
\label{sec:statistics}

We mine generics from a total of $50,534,844$ \textsc{Zyda} documents (23\% of the corpus). After the syntactic filtering for bare plurals, we end up with $16,771,049$ sentences, of which $\mathbf{4,146,760}$ make up the final \textsc{MGen} dataset after receiving a score of $0.8$ or higher by the generics classifier.

\paragraph{Source composition.} The final dataset contains over 3 million sentences from internet crawls (RefinedWeb, The Pile and SlimPajama) and around 1 million sentences from academic sources, peS2o and arXiv (Appendix \ref{apx:mgen}; Table \ref{tab:mgen_sources}). Of the total $4.1$ million samples, about $3$ million are bare plural generics, while the rest is made up of the $11$ quantifiers in different proportions (Table \ref{tab:updated_combined}).

\paragraph{Context documents.} For every sentence in \textsc{MGen}, we include the document from \textsc{Zyda} that contains it. These documents correspond to websites or papers and are generally long, averaging over 5000 words. For comparison, the context documents in the samples of \textsc{GenericsKB-Best} are much shorter, with an average of 147 words.

\begin{figure}
  \centering
  \includegraphics[width=\linewidth]{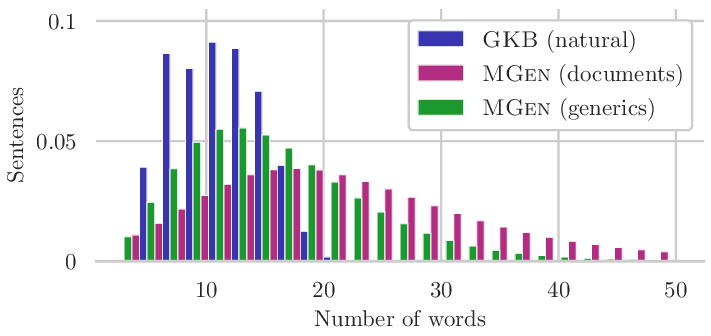}
  \caption{Sentence length distribution (percentage of samples for every number of words) in the generics and documents of \textsc{MGen} and natural sentences in \textsc{GenericsKB-Best}.}
  \label{fig:word_counts}
\end{figure}

\paragraph{Sentence length.} We compute the length of sentences in words by splitting sequences by whitespaces. Figure \ref{fig:word_counts} compares sentence length distributions for the naturally occurring examples in \textsc{GenericsKB-Best}, the generic (not quantified) sentences in \textsc{MGen} and the lengths in a sample of 20,000 context documents from \textsc{MGen} (Figure \ref{fig:word_counts}).

Generic sentences in \textsc{MGen} have an average of $16.65\pm8.2$ words and a median of $15$ words: generics are often long sentences. 
Although generics are on average shorter than arbitrary sentences from \textsc{MGen} documents, the length distribution 
contrasts with the prototypical examples in the linguistics and philosophy literature, as well as many synthetic examples in computational linguistics, that usually have less than 5 words (see Table \ref{tab:classic_generics} and examples in the Discussion \S\ref{sec:discussion}). Examples of sentences in \textsc{MGen} with lengths from 3 to 25 words are available in Table \ref{tab:sentence_length_examples} (Appendix \ref{apx:annotations}).


\paragraph{Common words.} The 50 most common words (excluding stopwords and punctuation) in \textsc{MGen} also reveal interesting aspects of the use of generics (Appendix \ref{apx:mgen}; Table \ref{tab:common_words}). 

The most common word in \textsc{MGen} generics is \textit{people}, with a big gap with respect to the second and third most common words: \textit{also} and \textit{cells}. In the generics of  \textsc{GenericsKB-Best}, \textit{also} is the most common word, and \textit{water} and \textit{one} are both more frequent than \textit{people}, which is still fourth. 

Following \textit{people}, \textit{women} and \textit{children} are nouns with many occurrences, as well as terms specific to biology and medicine, such as \textit{cells} and \textit{patients}. 
The most common verb is \textit{use} (and \textit{used}, from passive constructions).

In contrast, we analyze the most common words in $100,000$ context documents from \textsc{MGen} and find that \textit{people} does not even appear in the top 50: it is almost 60 times less prevalent ($16,5384$) than the most common word, which is \textit{also} with $942,208$ appearances.

These surface statistics of the sentences in the dataset 
give clues as to  how we use generic sentences: to generalize about \textit{people} and to express what to \textit{use} things for. 

In biology and medicine academic domains, which are well-represented in our dataset, we find a widespread use of generic sentences, as can be seen by the high frequency of some nouns particular to those fields.

\subsection{Human evaluation of \textsc{MGen}}
\label{sec:genericity}

To evaluate the quality of samples in the \textsc{MGen} dataset in terms of genericty we use human annotators.

We sample 300 sentences from \textsc{MGen} which get annotated by two annotators by labeling the sentences as \textit{Generic}, \textit{Particular} (non-generic) or \textit{Unclear}. Annotator guidelines are available in Appendix \ref{apx:annotation}. Examples with both annotations and the score of the \textsc{RoBERTa} classifier can be found in Table \ref{tab:annotation_disagreements} and Table \ref{tab:annotation_examples} (Appendix \ref{apx:annotations}). 

Annotators label $87.17\%$ sentences as \textit{Generic}, $7.5\%$ as \textit{Unclear} and $5.33\%$ as \textit{Particular}, with an $82\%$ of inter-annotator agreement. Table \ref{tab:annotation_disagreements} contains examples of disagreements. The human evaluation results suggest that, even as the annotation of generics is done automatically by a rather small model, the overall quality of the samples in \textsc{MGen} is high, making it a reliable source for generic sentences in context.





\begin{table*}[h]
  \centering
  \small
  \begin{tabular}{lcrrrccc} 
  \toprule
  & \multicolumn{1}{c}{diversity-from-similarity} & \multicolumn{3}{c}{distinct $n$-grams ($1M$ tokens)} & \multicolumn{3}{c}{head lemmas ($200k$ sentences)}  \\
  & $m_\texttt{cossim}$ & distinct-$1$ & distinct-$2$ & distinct-$3$ & Subject & Verb & Object  \\
  \midrule
  \textsc{MGen} & $\mathbf{-7.09\pm0.13}$ & $\mathbf{31,554}$ & $\mathbf{396,923}$ & $\mathbf{700,782}$& $\mathbf{18,836}$ & $\mathbf{7,131}$& $\mathbf{15,935}$   \\
  \textsc{GenericsKB} & $-8.27\pm0.14$ &$24,130$ & $308,320$ & $561,549$& $14,445$ & $5,133$ & $11,548$  \\
  \textsc{Gen-A-Tomic} & $-15.64\pm0.2$ & $19,398$ & $193,618$ & $357,334$ & $12,120$ & $3,909$ & $11,093$ \\
  \bottomrule
  \end{tabular}
  \caption{Diversity comparison of \textsc{MGen}, \textsc{GenericsKB-Best} and \textsc{Gen-A-Tomic}. In all scores higher is better.}
  \label{tab:diversity}
\end{table*}

\subsection{Diversity}
\label{sec:diversity}

We evaluate the diversity of the \textsc{MGen} dataset using three different measures: cosine similarity of sentence embeddings, distinct $n$-grams and distinct lemmas at subject, verb and object head positions.

\paragraph{Diversity from cosine similarity.} \citet{tevet_berant2021diversity} introduce a transformation from pairwise sentence similarity to a diversity metric by taking an average of the similarity across possible sentence pairs (Eq. \ref{eq:diversity_from_similarity}).

Given a corpus $\mathcal{C}$ and a 2-sentence similarity metric $m_\text{sim}(s_1,s_2)\in\mathbb{R}; s_1,s_2\in\mathcal{C}$, the corresponding diversity-from-similarity metric as:
\begin{equation}
  \label{eq:diversity_from_similarity}
  D_\text{sim}(\mathcal{C})=-\frac{1}{{{|\mathcal{C}|} \choose 2}}\sum_{s_i,s_j\in\mathcal{C}; i<j}m_\text{sim}(s_i,s_j)
\end{equation}
We use as similarity function the cosine similarity ($m_\texttt{cossim}$) between sentence embeddings generated with \textsc{NV-Embed-v2} \citep{lee2024nv}, a state-of-the-art model\footnote{As of December 2024.} in the Massive Text Embedding Benchmark \citep{muennighoff2023mteb}.

This diversity metric is computationally intractable for datasets with millions of sentences, we instead take 1000 samples of 1000 sentences each from the different datasets and report average diversity. 

\paragraph{Diversity in distinct $n$-grams.} We also consider an $n$-gram based diversity score, the distinct-$n$ score \citep{li2015diversity}.

Given a corpus $\mathcal{C}$ with $N_n$ $n$-grams and $U_n$ unique $n$-grams. Then, the \textit{distinct-$n$} score of $\mathcal{C}$ is the number of distinct $n$-grams ($U_n$) divided by the total number of words ($N_1$) in the corpus.
\begin{equation}
    \text{distinct-}n_\mathcal{C}=\frac{U_n}{N_1}
\end{equation}

We sample sentences from the each dataset until we reach $1$ million tokens (as per the \textsc{RoBERTa} tokenizer). For clarity, we report the number of distinct $n$-grams directly, without normalizing by $N_1$, as all samples have the same size in total tokens.

\paragraph{Diversity from head lemmas.} Because sentences in \textsc{MGen} are naturally occurring, samples may have relative, subordinated or conjunctive clauses beyond the main bare plural generic, which could artificially inflate the $n$-gram count. 

To have a fair comparison in this regard we introduce a score that counts the unique lemmatized verbs and head nouns in the subject and object positions. For each generic sentence, we get at most 3 lemmas, regardless of any clauses or subordinated sentences. For example, given \textit{bees in the forests of Catalonia feed on lavender flowers, giving their honey a distinctive taste} would be reduced to 3 lemmas: \textit{bee}, \textit{feed} and \textit{flower}. This way we target more directly the diversity in the generic sentences of the dataset.

We sample $200,000$ sentences from each dataset and report the total unique lemmas found.

\paragraph{\textsc{MGen} is the most diverse generics dataset.} We compare \textsc{MGen} to \textsc{GenericsKB-Best} and \textsc{Gen-A-Tomic} in terms of diversity by the three previous measures (Table \ref{tab:diversity}). To make the comparison fair, we leave out synthetic samples from \textsc{GenericsKB-Best}, and use only the naturally occurring sentences. 

In all cases, \textsc{MGen} is more diverse than the comparable datasets of generics, both in lexical (distinct $n$-grams and head lemmas) and neural (cosine similarity) measures. This shows that the \textsc{RoBERTa} classifier, even if it is based on a relatively small model, is able to label a wide range of generics. 

\section{Discussion}
\label{sec:discussion}

In recent years, the study of generic sentences has focused on the careful consideration of a series of prototypical examples that highlight different aspects of their semantics. Some notable generics are \textit{typhoons arise in this part of the Pacific}{ }\citep{carlson1977disseration}, \textit{mosquitoes carry the West Nile virus}{ }\citep{generics_08_leslie}, \textit{ducks lay eggs}{ }\citep{leslie2011gog}, \textit{humans kill themselves}{ }\citep{sterken2015context}, \textit{dobermans have floppy ears}{ }\citep{nickel2016between} and many others. Although these examples are effective at illustrating the semantics of generics, they are difficult to leverage computationally.

With the introduction of \textsc{MGen}, a massive collection of naturally occurring generics in context, 
we open the door for new computational and corpus-level approaches to make progress in the puzzle of generics.

\textsc{MGen} consists of 3 million generics and 1 million sentences explicitly quantified by 11 different quantifiers. These have been mined from a diverse pool of internet and academic documents, ensuring that many of the ways in which speakers use generics are represented.

Our analysis shows that \textsc{MGen} is the more diverse of the large-scale datasets of generics, and human annotation suggests that, even as generics are automatically filtered, the quality of the examples is high.


If we take \textsc{MGen} as a representative sample of generics, at least of some of the many ways in which English speakers use them, the statistics of the dataset say much about generics themselves.

The analysis of sentences in \textsc{MGen} suggests that \textit{generics are long}. They have over 16 words on average, with the most common sentence length being 15. Even if some generics in the dataset are long due to clauses and subordinate sentences, this still suggest sentences that begin with a generic express complex ideas.
We also find many generics, in scientific and medical domains \citep{peters2024medicalgenerics}, that are not only long but contain many technical terms. 

The technicality and length of many generics in \textsc{MGen} contrasts with theories that link generics to "thinking-fast" or System I \citep{kahneman2011fastslow} in the dual-process theory of cognition \citep{leslie2007structure, gencontext_23_almotahari}. Combining the intuitive and unreflective use of generics, which speakers often do, with some of the long and complex sentences in \textsc{MGen} is one of the open questions this dataset could help resolve.

We believe \textsc{MGen} can play a role in future research on generics and quantifiers by providing examples with long context documents across a multiple sentence lengths (Appenix \ref{apx:annotations}; Table \ref{tab:sentence_length_examples}) and topics, from academic papers to internet forums. These could disclose different ways in which speakers use generics. For example, that \textit{people} is the most common noun suggests that generics play an important role on how humans understand each other through language.


\section{Conclusion}

In this work we build \textsc{MGen}, a massive collection of generic and quantified sentences in context.

We mine generic sentences from \textsc{Zyda}, a corpus for language model training. Our two-step pipeline first filters sentences by their syntactic features and then uses a \textsc{RoBERTa}-based classifier to determine genericity.

The final dataset contains over 3 million bare plural generics and 1 million quantified sentences with 11 different quantifiers. We believe \textsc{MGen} is a valuable resource for future research on generic sentences.

The \textsc{MGen} dataset is open-source, available at \href{https://gustavocilleruelo.com/mgen}{\tt gustavocilleruelo.com/mgen}.

\section*{Acknowledgements}

This work was funded by UK Research and Innovation (UKRI) under the UK government’s
Horizon Europe funding guarantee [grant number
10039436].

The first author would like to thank Mahrad Almotahari, Dan Lassiter, Nicolas Navarre and Aina Centelles Tarrés for their proof-reading, corrections and discussions.

We also thank the anonymous reviewers for their
insightful comments and feedback on the manuscript.

\section*{Limitations}

\paragraph{Data contamination.} This dataset is designed as a corpus for the study of language, rather than for any evaluation of the performance of language models. The sources that conform \textsc{Zyda} are commonly used in the training of language models, which means any sort of performance evaluation in this data would be compromised and should be carefully carried out.

\paragraph{Generics classifier.} The classifier that we use to classify generics as such does only take information from the sentence itself, we do not append any context. Future versions of the pipeline could use stronger models for selection of generics from bare plural sentences.

\paragraph{Distribution of generics.} Although \textsc{MGen} has millions of generics, it may not capture the full distribution of generic sentences: it only contains bare plural generics at the beginning of the sentence. Similarly, the quantified sentences we select are within a limited range of structures.

Three main assumptions underlie the generics of this dataset: (i) bare plurals (ii) at the beginning of the sentence (iii) in English. Future work that tries to capture more holistically generics across languages should improve upon these.


\appendix

\renewcommand{\thefigure}{\thesection.\arabic{figure}}
\renewcommand{\thetable}{\thesection.\arabic{table}}
\setcounter{figure}{0}  
\setcounter{table}{0}  

\section{Training and evaluation of the generics classifier}
\label{apx:roberta}

\paragraph{Training.} We build the generics classifier by training a first iteration on generics from \textsc{GenericsKB} and then refining it iteratively. We make the training set more complete by adding examples the classifier struggles on from the candidate bare plurals, thus covering difficult and corner cases. We synthetically augment this challenging datapoints with the prompts in Appendix \ref{apx:synthetic_eval_prompts}. Table \ref{tab:classifiers} shows the final distribution of the training dataset, which trains a classifier that reaches $0.97$ F-$1$ score in our $3622$ sentences evaluation set.

\begin{table}[h]
  \centering
  \begin{tabular}{l c}
      \toprule
      {\bf Origin} & {\bf Sentences} \\
      \midrule
      \textsc{GenericsKB} (generics) & 2500 \\
      Synthetic non-generics & 2039 \\
      Non-generics from data & 310 \\
      Generics from data & 61 \\
      \bottomrule
  \end{tabular}
  \caption{Composition of the \textsc{RoBERTa} classifier training data.}
  \label{tab:classifiers}
\end{table}

\paragraph{Evaluation data.} We evaluate the generics classifiers in \textsc{ConGen} for positive examples and a synthetic negative examples generated with \textsc{GPT-4} \citep{openai2024gpt4technicalreport}. We include the quantified sentences in \textsc{ConGen} by removing the quantifier (\textit{most tigers hunt rabbits} becomes \textit{tigers hunt rabbits}). The negative (non-generic) sentences are designed to be challenging for a generics classifier (details are available in Appendix \ref{apx:synthetic_eval_prompts}). The final test set includes 3622 test sentences: 2873 generics and 749 non-generics. 

\section{Synthetic adversarial non-generic bare plurals generation}
\label{apx:synthetic_eval_prompts}

We combine variations of the following prompts to generate synthetic data based on difficult examples in the data, where iterations of the generics classifier struggle. We also focus on filtering out some examples undetectable to the synthetic filtering step, such as sentences with the title section present (for example, \textit{Introduction Transformers are function approximators}). We use some of the synthetic examples generated for the training and some for the evaluation of the classifier.

\paragraph{Prompt\#1.} \texttt{Task: generation of declarative sentences indicative that are not generic. The sentences generated should not be generic sentences, even if they share features with them. The following examples are non-generic sentences, or sentences that do not begin with the generic sentence. \\ \\ Examples: \\\{ list of examples\}\\ \\Based on the previous examples, generate 100 non-generic sentences using a wide range of vocabulary and basing the generated sentences on the types of syntax in the examples, and other varied syntactic constructions similar to bare plurals, such as adding elements that make it so that the generic sentence is not at the beginning or is not grammatical. The setences cannot begin with a generic, such as "tigers have stripes" or "nerves carry messages throughout the body", but rather existentials, ungrammatical or beginning with a section title. Generate the examples in the format of a python list of strings.}

\paragraph{Prompt\#2.} \texttt{Task: generate existential sentences that syntactically resemble bare plural generic sentences. For examples are sentences that talk about figures, equations, examples and studies in scientific articles, such as "Blue arrows indicate acceleration", "Examples of this are equations 2 and 4" or "Studies show this phenomena happens often". Can you generate 100 sentences like these in a python list sentence. Make them with varied lengths and lexically varied, and make sure they are clearly not generic, for example by referencing figure numbers etc.}

\paragraph{Prompt\#3.} \texttt{Generate 10 sentences that have a similar structure than the following example. Return the results in the format of a python list. \\ \\ Example: Processes are made of repetitive...}

\section{Sentence Length in \textsc{Mgen}}
\label{apx:sentence_length}

The 20,000 sampled documents sampled from \textsc{MGen} yield a total of 4,202,451 sentences.

\begin{table}[h]
  \centering\small
  \begin{tabular}{l r r}
      \toprule
      {\bf Dataset} & {\bf Average } & {\bf Median } \\
      \midrule
      \textsc{MGen} (generics) & $16.65\pm8.2$ & $15$ \\
      \textsc{Mgen} (documents) &$24.75{\pm29.3}$&$21$ \\
      \textsc{GenericsKB-Best} (natural) & $9.66\pm3.66$ & $10$ \\
      \bottomrule
  \end{tabular}
  \caption{Average and median length across datasets.}
  \label{tab:avg_length}
\end{table}

\section{Annotation of \textsc{MGen}}
\label{apx:annotation}

These are the instructions and examples annotators received:
\begin{itemize}
  \item[·] Assign the label “Generic”, “Particular” or “Unclear” to each sentence in your sheet.
  \item[·] “Generic” sentences make a broad statement that applies to members of a category or group in general. For example, \textit{Birds fly}, \textit{German shepherds are loyal}, \textit{Well-maintained public parks attract visitors all year-round}. Even if the group is very specific, such as \textit{Red birds with long beaks that live in the jungle fly}, as long as it does not appear like the text refers to specific individuals in the context, label it as a generic.
  \item[·] “Particular” sentences talk about a specific set of individuals or events. They usually provide information about one or a few individuals in a group: \textit{This bird can fly}, \textit{Dogs are in the front lawn}. These are sentences that talk about particular things in a context: \textit{Units are in kilograms}, \textit{Arrows indicate acceleration} would not be generics as they only make sense when refering to a specific table or plot. \textit{German shepherds outside the house are loyal} is also not a generic, as it refers to specific german shepherds.
  \item[·] In case of subsentences, focus only on the first subsentence: \textit{Birds fly and this parrot speaks} would still count as generic even if "this parrot speaks" is not a generic since it refers to a particular parrot.
  \item[·] Do not worry if you are unsure about whether a sentence is “Generic” or “Particular”. In this case, or if the sentence is grammatically incorrect, please use the “Unclear” label. Use also "Unclear" if you are not sure, you would need more context to answer or if the first words in the sentence are not a generic (for example: \textit{In any case, birds fly})
  \item[·] For more examples, have a look at the annotated sentences in red. Thank you for your participation!
\end{itemize}

They also had the following examples:

\begin{itemize}
  \item[·] Tigers have stripes. \textit{Generic}
  \item[·] Tigers have stripes, they are cats and the ones we have here are violent. \textit{Generic}
  \item[·] Those tigers have stripes. \textit{Particular}
  \item[·] Tigers, which are part of the Felidae family, have stripes. \textit{Generic}
  \item[·] Tigers in this zoo are violent. \textit{Particular}
  \item[·] Tigers in zoos are violent. \textit{Generic}
  \item[·] Tigers are in the front lawn. \textit{Particular}
  \item[·] Tigers are also like this. \textit{Generic}
  \item[·] Tigers share that characteristic with lions. \textit{Generic} 
\end{itemize}

\begin{figure}[H]
  \centering
  \includegraphics[width=\linewidth]{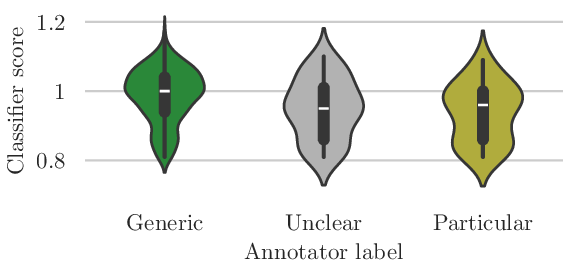}
  \caption{Correspondence of human annotations with \textsc{RoBERTa} classifier scores.}
  \label{fig:annotation_labels}
\end{figure}

\section{Composition of the \textsc{MGen} dataset}
\label{apx:mgen}

Table \ref{tab:zyda_components} shows the millions of documents each component of \textsc{Zyda} has. Note that we only mine generics from about $23\%$ of the dataset. The final amount of sentences in \textsc{MGen} by source is in Table \ref{tab:mgen_sources}.

Finally, Table \ref{tab:common_words} shows the top 50 common words for generics in \textsc{MGen}, naturally occurring sentences in \textsc{GenericsKB-Best} and $100,000$ documents sampled from the contexts in \textsc{MGen}.

\begin{table}[h!]
  \centering
  \begin{tabular}{l c c}
      \toprule
      {\bf Source} & {\bf Total Documents} & {\bf Origin} \\
      \midrule
      RefinedWeb & $920.5M$ & Internet \\
      SlimPajama & $142.3M$ & Internet \\
      The Pile  & $64.9M$ & Varied\\
      peS2o & $35.7M$ & Academic \\
      arXiv & $0.3M$ & Academic \\
      \bottomrule
  \end{tabular}
  \caption{Information on the components of \textsc{Zyda} we run the generics pipeline on.}
  \label{tab:zyda_components}
\end{table}

\begin{table}[h]
  \centering
  \begin{tabular}{l r}
      \toprule
      {\bf Source} & {\bf Sentences} \\
      \midrule
      RefinedWeb  &          $1,270,280$ \\
      The Pile &   $1,019,687$ \\
      SlimPajama &           $993,373$ \\
      peS2o &                 $796,334$ \\
      arXiv &                 $67,086$ \\
      \bottomrule
  \end{tabular}
  \caption{Combined statistics for \textsc{MGen} by source.}
  \label{tab:mgen_sources}
\end{table}

\begin{table*}[h]
  \small
  \centering
  \setlength{\tabcolsep}{20pt}
  \begin{tabular}{lrlrlr}
  \toprule
    \multicolumn{2}{c}{\textsc{MGen} (generics)} & \multicolumn{2}{c}{\textsc{GenericsKB-Best}} & \multicolumn{2}{c}{\textsc{MGen} ($100k$ documents)} \\ 
    \textbf{Word} & \textbf{Count} & \textbf{Word} & \textbf{Count} & \textbf{Word} & \textbf{Count}\\
    \midrule
    people   & 200946 & also      & 23933 & also      & 942208 \\
    also     & 183012 & water     & 20301 & data      & 879361 \\
    cells    & 96700  & one       & 18145 & using     & 780702 \\
    used     & 96104  & people    & 16598 & one       & 767704 \\
    different& 94097  & many      & 12452 & model     & 735504 \\
    use      & 92326  & important & 12417 & used      & 727311 \\
    like     & 89778  & life      & 11283 & two       & 653421 \\
    one      & 84314  & plants    & 10967 & different & 591577 \\
    make     & 74173  & cause     & 10933 & figure    & 587311 \\
    high     & 70107  & common    & 10923 & time      & 585129 \\
    many     & 70083  & used      & 10715 & study     & 584773 \\
    need     & 70010  & body      & 10344 & results   & 576442 \\
    women    & 68460  & use       & 10074 & may       & 568490 \\
    time     & 64141  & different & 10036 & cells     & 539390 \\
    children & 61270  & food      & 9964  & al.       & 535876 \\
    well     & 60362  & animals   & 9315  & however   & 477362 \\
    systems  & 60005  & energy    & 8891  & use       & 476105 \\
    tend     & 57323  & human     & 8886  & number    & 474336 \\
    important& 56710  & cells     & 8858  & system    & 468788 \\
    provide  & 56523  & form      & 8660  & analysis  & 446709 \\
    work     & 55676  & time      & 8478  & first     & 445497 \\
    less     & 50941  & children  & 7757  & fig       & 438667 \\
    good     & 50521  & women     & 7618  & based     & 385968 \\
    much     & 48714  & blood     & 7147  & models    & 373924 \\
    get      & 47917  & light     & 7109  & high      & 372224 \\
    large    & 47588  & small     & 7086  & function  & 371581 \\
    small    & 47149  & disease   & 6953  & learning  & 370877 \\
    water    & 46181  & world     & 6884  & information& 370467 \\
    way      & 45507  & cancer    & 6653  & case      & 356658 \\
    even     & 44487  & natural   & 6583  & set       & 351422 \\
    common   & 44330  & like      & 6527  & shown     & 349042 \\
    may      & 43538  & part      & 6452  & table     & 348287 \\
    patients & 43443  & often     & 6257  & cell      & 341799 \\
    likely   & 43303  & large     & 6220  & new       & 334611 \\
    higher   & 43208  & make      & 6199  & given     & 330825 \\
    health   & 42758  & high      & 6148  & well      & 326821 \\
    help     & 41548  & air       & 6017  & studies   & 325837 \\
    men      & 40689  & health    & 5982  & patients  & 325434 \\
    system   & 40548  & live      & 5889  & research  & 321275 \\
    known    & 40036  & two       & 5774  & found     & 319645 \\
    play     & 39813  & way       & 5503  & could     & 317444 \\
    two      & 38604  & well      & 5478  & due       & 314760 \\
    human    & 38571  & means     & 5464  & see       & 312387 \\
    life     & 38428  & occurs    & 5447  & systems   & 306782 \\
    data     & 37663  & process   & 5403  & energy    & 304915 \\
    great    & 37612  & soil      & 5397  & thus      & 303428 \\
    form     & 37517  & occur     & 5373  & method    & 299352 \\
    new      & 37113  & growth    & 5157  & process   & 298258 \\
    n't      & 36267  & work      & 5145  & group     & 290830 \\
    social   & 36212  & system    & 5046  & would     & 289965 \\
    \bottomrule
  \end{tabular}
  \caption{Top 50 common words in generic sentences from \textsc{MGen} and \textsc{GenericsKB-Best}.}
  \label{tab:common_words}
\end{table*}

\section{Data samples}
\label{apx:annotations}

\begin{table*}[h]
  \centering
  \small
  \begin{tabular}{p{10cm}l}
  \toprule
  
  \textbf{{Bare plural}} & \textbf{{Source}} \\
  \midrule
  Solid lines are the analytical results (Eqs.&arXiv\\[0.2cm]

  State police report 30 year old Kira Zink was headed south \dots&SlimPajama\\[0.2cm]

  Svp binding sites are underlined.&The Pile\\[0.2cm]

  COST: Entries start at \$10; MORE INFO TUESDAY, DECEMBER 24\dots&SlimPajama\\[0.2cm]

  Online master's programs close on May 5th and August 19th.&SlimPajama\\[0.2cm]

  Tickets cost £12 (students £5, under 18s go free)\dots&RefinedWeb\\[0.2cm]
  
  \bottomrule
  \end{tabular}
  \caption{Examples of existential (non-generic) bare plurals from \textsc{Zyda}. Dots (\dots) indicate the example was truncated.}
  \label{tab:existentials}
\end{table*}

\begin{table*}[h]
  \centering
  \small
  \begin{tabular}{p{10cm}l}
  \toprule
  
  \textbf{{Sentences}} & \textbf{{Source}} \\
  \midrule
  Horses are mammals &\citep{carlson1977unified}\\[0.2cm]
  Horses are larger than mules &\citep{carlson1977unified}\\[0.2cm]
  Elephants are easily trained &\citep{carlson1977unified}\\[0.2cm]
  Mosquitoes carry the West Nile virus &\citep{generics_08_leslie}\\[0.2cm]
  Cats have whiskers &\citep{generics_08_leslie}\\[0.2cm]
  Peacocks have fabulous blue tails &\citep{generics_08_leslie}\\[0.2cm]
  Diamonds are valuable & \citep{nickel2016between}\\[0.2cm]
  Elephants live in Africa or Asia & \citep{nickel2016between}\\[0.2cm]
  Coke bottles have short necks & \citep{nickel2016between}\\[0.2cm]
  Cabs are yellow & \citep{sterken2015context}\\[0.2cm]
  Birds lay eggs, but mammals don’t. Mammals give birth to live young. & \citep{sterken2015context}\\[0.2cm]
  Lottery tickets are losers & \citep{sterken2015context}\\[0.2cm]

  \bottomrule
  \end{tabular}
  \caption{Some generics that serve as running examples in the literature.}
  \label{tab:classic_generics}
\end{table*}

\begin{table*}[h]
  \small
  \centering
  \begin{tabular}{p{10cm} c c c}
    \textbf{Text} & \textbf{Label 1} & \textbf{Label 2} & \textbf{Score} \\
    \midrule
    Textbooks provide templates for proper procedure: the who, why, what, and where of the story. & Generic & Generic & 0.91 \\
    Flatforms are comfy because of the uniform thickness of the heel and at the same time practical and easy to style in the morning with jeans and T-shirts and in the evening with Oversized Dresses. & Generic & Generic & 0.90 \\
    Males have two sex organs, known as hemipenes, which are normally kept within the body, but are everted from his vent for mating. & Unclear & Generic & 1.06 \\
    Cash crops are called commercial or commercial crops. & Generic & Generic & 1.03 \\
    Oil-based primers are also very good remedies for covering staining on walls and ceilings that have oil-based paints. & Generic & Generic & 1.02 \\
    Thin clients are less intelligent terminals that connect to applications hosted on a remote computer. & Unclear & Generic & 1.03 \\
    Thicker greens such as romaine or bib lettuce are better for salads that will have a lot of meat or chunky vegetables. & Generic & Generic & 1.07 \\
    JWs today have a similar command structure to promote uniformity rather than truth and love, in every element of a Christians life. & Generic & Generic & 0.95 \\
    People realize that the best way to control their housing costs is ownership. & Generic & Generic & 1.03 \\
    People who wish to argue against Spiritualism are quite sure, as a rule, that media will descend to any trickery and cheating for the sake of gain. & Generic & Generic & 0.93 \\
    Red d'Anjou pears are excellent for fresh eating, poaching, cooking and all types of baking. & Generic & Generic & 0.95 \\
    Powerful computing systems also require high speed access to large data storage systems. & Generic & Generic & 0.95 \\
    Filipinos of Hispanic ancestry form a minority in the Philippine population. & Generic & Generic & 1.06 \\
    IMTs operate in various ways. & Generic & Unclear & 0.99 \\
    Weak institutions lead to weak coordination and fragmented interventions that often prove ineffective. & Generic & Generic & 1.04 \\
    Ventilation flaps are used in the air ducts of heating and ventilation systems or air conditioning systems in an automobile and are usually adjusted via Bowden pull mechanisms or mechanical transmissions. & Generic & Generic & 1.05 \\
    Quantum computers promise to directly simulate systems governed by quantum principles, such as molecules or materials, since the quantum bits themselves are quantum objects. & Generic & Generic & 1.04 \\
    Pair bonds are monogamous and seasonal. 3–6 eggs are incubated by the female only, but the chicks are usually brooded and fed by both birds. & Generic & Generic & 1.03 \\
    Puppets are fun to include too. & Particular & Unclear & 0.86 \\
    Parenchyma cells are also responsible for healing in the plant - this tissue can go through cell division and regenerate when needed. & Generic & Generic & 1.03 \\
    Conventional linear synchronous motors have issues of high manufacturing cost of the stator and high magnetic loss. & Generic & Generic & 0.99 \\
    Traditions are a vital a part of the Italian culture and naturally, weddings have their very own. & Generic & Unclear & 0.92 \\
    Calm dog breeds include Great Danes, Great Pyrenees, Basset Hounds, Shih Tzus, and Pugs. & Unclear & Unclear & 0.84 \\
    First thoughts are proverbially the best; at all events, they are the bravest. & Unclear & Generic & 0.96 \\
    Bursts are by definition variable, as temperature evolution due to thermonuclear burning and then cooling drives the fast increase and then slower decrease in X-ray flux. & Particular & Generic & 0.97 \\
    People are under pressure to make the systems efficient, but they are expected to keep the system safe, which inevitably introduces inefficiencies. & Particular & Generic & 0.91 \\
    Police officers are human beings, and many of them understand that the pressures of everyday life can sometimes lead good drivers to make bad decisions. & Generic & Generic & 1.11 \\
    Self-induction habits are oft described as a compulsive behavior, with magnetic-like attraction to light sources commonly reported [9]. & Generic & Generic & 0.88 \\
    Gastroenterologists, infectious disease specialists, hepatologists, and even some nurse practitioners commonly manage cases of Hep C. & Unclear & Generic & 1.1 \\
    Natural degradable polymers and their composites are amongst these materials. & Particular & Generic & 0.84 \\
    Involving surrounding tissue structures, tonsillar tumours often infiltrate the soft palate, the base of the tongue, the lateral pharyngeal wall and medially the parapharyngeal space as well as the vascular sheath. & Generic & Unclear & 0.83 \\
    Caries are understood to result from the accumulation of plaque on the teeth and the production of organic acids (plaque acids) when plaque microorganisms ferment sugars and starches in food. & Generic & Generic & 1.06 \\
    Female beetles deposit their eggs singly on the legume seeds. & Generic & Generic & 1.06 \\
      \bottomrule
  \end{tabular}
  \caption{33 examples from \textsc{MGen} generics with both annotations and scores.}
  \label{tab:annotation_examples}
\end{table*}

\begin{table*}
\centering
\small
\begin{tabular}{cp{0.5\textwidth}cc}

\textbf{Length} & \textbf{{Generic}} & \textbf{{Source}} & \textbf{{Score}} \\
\toprule
3 & Words have power. & RefinedWeb & 0.98 \\[0.2cm]
4 & Democrats are control freaks. & The Pile & 1.01 \\[0.2cm]
5 & Children learn what they live. & The Pile & 1.08 \\[0.2cm]
6 & Ghosts represent a post-death human consciousness. & SlimPajama & 1.02 \\[0.2cm]
7 & Color and pictures are fun and vibrant. & RefinedWeb & 0.82 \\[0.2cm]
8 & More complex bytecodes trap to a software routine. & peS2o & 0.85 \\[0.2cm]
9 & Males tend to be more affected by the disease. & SlimPajama & 0.99 \\[0.2cm]
10 & Triggers cause individuals to become ineffective and produce negative energy. & The Pile & 1.02 \\[0.2cm]
11 & Professional massage therapists relieve tired muscles and alleviate pain in customers. & RefinedWeb & 0.97 \\[0.2cm]
12 & American workers produce sophisticated goods or investment opportunities at lower opportunity costs. & SlimPajama & 1.06 \\[0.2cm]
13 & Insurance companies reward property owners who personal their house totally free and obvious. & RefinedWeb & 1.0 \\[0.2cm]
14 & Alkaline phosphatases carry out hydrolase/transferase reactions on phosphate-containing substrates at a high pH optimum. & The Pile & 1.0 \\[0.2cm]
15 & Stimulants are substances that raise the levels of physiological or nervous activity in the body. & RefinedWen & 1.04 \\[0.2cm]
16 & Areas along large rivers are commonly inhabited by baldcypress, water tupelo, water elm, and bitter pecan. & The Pile & 0.94 \\[0.2cm]
17 & Sports fans are far more familiar with NBC Sports, which televises everything from Super Bowls to Olympics. & The Pile & 0.96 \\[0.2cm]
18 & Keto dieters love exogenous ketones because they help fight the keto flu and get you quickly into ketosis. & The Pile & 1.07 \\[0.2cm]
19 & Insects evolve adaptations allowing them to eat specific species of plants, while being unable to eat most other plants. & RefinedWeb & 1.04 \\[0.2cm]
20 & Extractive methods, such as lipoplasty (liposuction) or local excision, are methods whereby fat is mechanically removed from areas of interest. & The Pile & 0.96 \\[0.2cm]
21 & Factory-terminated systems are also the only viable solution to the extremely low-loss systems that are required to support high-speed optic links. & RefinedWeb & 0.86 \\[0.2cm]
22 & Small Business consultants typically develop relationships with their customers and often correspond by e-mail with their customers and return customers' phone calls. & The Pile & 0.99 \\[0.2cm]
23 & Initial parton showers interact with the medium via collisional and radiative processes that cause dissipation and redistribution of energy inside the parton shower. & peS2o & 0.93 \\[0.2cm]
24 & Green superfoods have the highest concentrations of simply digestible nutrients, fat burning compounds, nutritional vitamins and minerals to safeguard and mend your body. ! & RefinedWeb & 0.87 \\[0.2cm]
25 & Punitive damages are awarded to punish a defendant for particularly egregious conduct, and to serve as a deterrent to future conduct of the same type. & The Pile & 0.96 \\
\bottomrule
\end{tabular}
\caption{Examples of generics from \textsc{MGen} at different sentence lengths.}
\label{tab:sentence_length_examples}
\end{table*}

\end{document}